# *An Efficient Color Face Verification Based on 2-Directional 2-Dimensional Feature Extraction*

**Lan-Ting LI**

Shanghai Institute of Applied Physics, Chinese Academy of Sciences

**Abstract**  A novel and uniform framework for face verification is presented in this paper. First of all, a 2-directional 2-dimensional feature extraction method is adopted to extract client-specific template — 2D discrimant projection matrix. Then the face skin color information is utilized as an additive feature to enhance decision making strategy that makes use of not only 2D grey feature but also 2D skin color feature. A fusion decision of both is applied to experiment the performance on the XM2VTS database according to Lausanne protocol. Experimental results show that the framework achieves high verification accuracy and verification speed.

## 1  Introduction

Face verification, compared with other biometric like signature, iris or fingerprints verification, has many advantages, i.e. non-intrusion and free-contact. Face verification is a special face recognition, which only need make a 2-class decision: accept as a client or reject as an imposter. So many typical face recognition methods can be applied to face verification, such as artificial neural networks (ANN, Georg Dorffner, Horst Bischof, Kurt Horn, 2001), support vector machines (SVMs, G. Guo, S.Z. Li, K. Chan, 2000), Fisher's linear discriminant (FLD, Belhumeur et al., 1997), etc.

In several face verification application scenarios, a low-complexity algorithm, which can be implemented on a low-cost processor, is desirable. Some examples of this situation are mobile phone, PDA, smart card or stand-alone control access systems. ANN and SVMs cannot adapt to the application scenarios, because of their high computation complexity. FLD is a low-complexity algorithm, but its verification accuracy is not satisfying. Therefore, many researchers proposed some algorithms to improve the verification accuracy of FLD [1, 2]. The typical example is that Josef Kittler and Yongping Li presented client specific fisherfaces for face verification (CS-LDA) [2], in which each registered client has its own fisher's projection vector.

In this paper, a novel framework, which extends features extracted to 2-dimension space, is used to overcome the disadvantages above. Moreover, skin color feature is helpful to face verification [3]. The grey information and color information are combined as face features. This paper is organized as follows. In section 2, how to construct a 2D space feature with 2-directional 2-dimensional $((2D)^2)$ feature extraction is introduced. Section 3 provides a feature model of face skin color information. Section 4 describes the experiments, in which many algorithms are compared in two aspects: verification rate and speed. Finally, conclusions are drawn in section 5.

## 2  $(2D)^2$ feature extraction

Daoqiang Zhang and Zhihua Zhou proposed $(2D)^2$PCA in 2005, the experiments proved $(2D)^2$PCA can achieve higher recognition rate and speed than PCA [4]. P.Nagabhushan et al. presented $(2D)^2$FLD in 2006, the experiments proved $(2D)^2$FLD can also achieve a better performance [5].

In this paper, the two methods are fused to be an effective $(2D)^2$ feature extraction algorithm. Suppose that the ith client has $n_i$ samples for training. The total client training samples is N. Consider an $m \times n$ client sample matrix A. Let $X \in R^{n \times d}$ be a matrix with orthonormal columns ($n > d$), and then $Z \in R^{q \times m}$ be a matrix with orthonormal rows ($m > q$). Projecting A onto X and Z yields an $q \times d$ matrix

$$Y = ZAX \qquad (1)$$

The column direction total scatter matrix $SC_T \in R^{n \times n}$ can be evaluated by

$$SC_T = \frac{1}{N}\sum_{i=1}^{N}(A_i - \overline{A})^T(A_i - \overline{A}) \quad (2)$$

where $A_i$ is the ith training sample and $\overline{A}$ denotes the average matrix of the total training clients samples.

The optimal column direction projection matrix $X_P \in R^{n \times g}$ can be obtained by computing the eigenvectors of $SC_T$ corresponding to the g largest eigenvalues.

Similarly, the row direction total scatter matrix $SR_T \in R^{m \times m}$ can be written as

$$SR_T = \frac{1}{N}\sum_{i=1}^{N}(A_i - \overline{A})(A_i - \overline{A})^T \quad (3)$$

The optimal row direction projection matrix $Z_P \in R^{h \times m}$ can be obtained by computing the eigenvectors of $SR_T$ corresponding to the h largest eigenvalues.

So the projected feature matrix $W \in R^{h \times g}$ will be obtained as

$$W = Z_P A X_P \quad (4)$$

Let $I_j^k \in R^{h \times g}$ represents the kth projected feature matrix in the jth class. Face verification is a 2-class recognition problem.

For example, if the ith client as claimed identity is Class one, the others clients as imposters will be Class two, then the column direction within-class scatter matrix $SC_W \in R^{g \times g}$ can be evaluated by

$$SC_W = \frac{1}{N}\left(\sum_{k=1}^{n_i}(I_i^k - M_c)^T(I_i^k - M_c) + \sum_{j=1, j\neq i}^{D-1}\sum_{k=1}^{n_j}(I_j^k - M_I)^T(I_j^k - M_I)\right) \quad (5)$$

where D is the number of all clients, $M_c = \frac{1}{n_i}\sum_{k=1}^{n_i} I_i^k$ denotes the ith client average matrix and $M_I = \frac{1}{N-n_i}\sum_{j=1, j\neq i}^{D-1}\sum_{k=1}^{n_j} I_j^k$ denotes imposters average matrix.

The column direction between-class scatter matrix $SC_B \in R^{g \times g}$ can be evaluated by

$$SC_B = (M_c - M_I)^T(M_c - M_I) \quad (6)$$

The optimal column direction projection matrix $X_F \in R^{g \times d}$ can be obtained by computing the eigenvectors of $SC_W^{-1} SC_B$ corresponding to the d largest eigenvalues.

According to Eq. (5),

$$rank(SC_W) \leq (N - D) \cdot \min(h, g) \quad (7)$$

So $SC_W$ is nonsingular when

$$N \geq D + \frac{g}{\min(h, g)} \quad (8)$$

In real situation, Eq. (8) is always satisfied, which is different from FLD.

Similarly, the row direction within-class scatter matrix $SR_W \in R^{h \times h}$ can be given by

$$SR_W = \frac{1}{N}\left(\sum_{k=1}^{n_i}(I_j^k - M_c)(I_i^k - M_c)^T + \sum_{j=1, j\neq i}^{D-1}\sum_{k=1}^{n_j}(I_j^k - M_I)(I_j^k - M_I)^T\right) \quad (9)$$

The column direction between-class scatter matrix $SR_B \in R^{h \times h}$ can be evaluated by

$$SR_B = (M_c - M_I)(M_c - M_I)^T \quad (10)$$

The optimal column direction projection matrix $Z_F \in R^{q \times h}$ can be obtained by computing the eigenvectors of $SR_W^{-1} SR_B$ corresponding to the q largest eigenvalues.

In real situation, $SR_W$ is also always nonsingular.

So the projected feature matrix $Y \in R^{q \times d}$ will be obtained by:

$$Y = Z_F W X_F = Z_F Z_P A X_P X_F \quad (11)$$

According to Eq. (1), $Z = Z_F Z_P$ and $X = X_F X_P$.

In Eq. (11), Y is a matrix of the size q × d. However, for CS-LDA [2], the projected result is a number. For the proposed framework, the performance of linear separability between client and imposter is better.

## 3 Skin color feature model

Many published studies have described that the decision level fusion of experts utilize multispectral information can lead to performance improvement [6], such as skin color information. Faces often have a characteristic color which is possible to separate from the others. According to

[7], in the color space YCrCb (Y denotes the grey component, Cr and Cb denotes the chroma components), Cr and Cb are sensitive to the variety of skin color. In addition, Cr and Cb obey Gaussian distribution, as showed in Fig. 1.

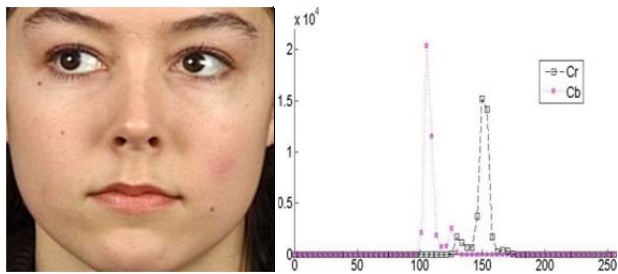

**Fig. 1** The histograms of Y, Cr and Cb

Two persons are selected from the face database XM2VTS arbitrarily showed in Fig. 2(a), each person have two face images both of which are shot at different periods. In real situation, the use of opponent chroma-CrCb (Cr − Cb) would yield more robust results. The face images constructed by (Cr − Cb) are described Fig. 2(b). The histograms of the face images are showed in Fig. 2(c).

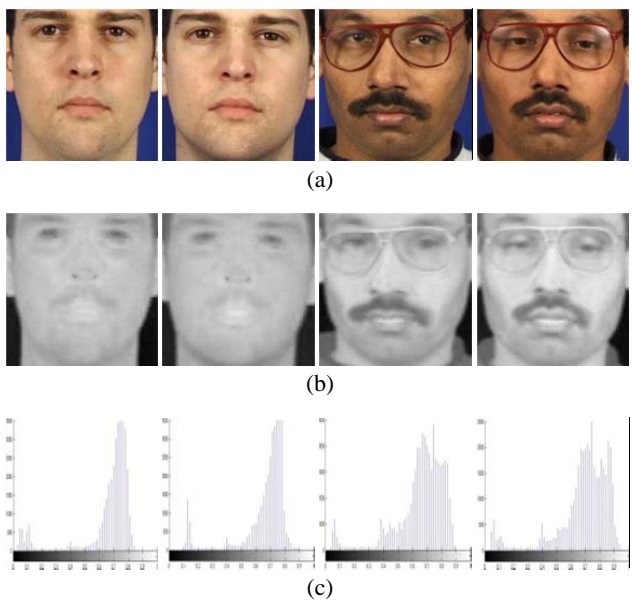

**Fig. 2** (a) The original face images (Note: The face images are captured on the same even illumination.); (b) The face images constructed by (Cr-Cb); (c) The histograms of the images in (b).

A conclusion can be drawn from Fig. 2, the histograms are characteristic for a specific person, but are also discriminant among different persons.

However, there is a weak point in the method. The color is similar between hair pixels and skin pixels. So the pixels in sub-windows previously selected should have the hair pixels as few as possible. Two measures can be applied: Firstly, in the stage of face detection, standard frontal face will be located without hair; secondly, skin color detection is used to eliminate the effect of hair pixel.

## 4 Experiments

### 4.1 The database and protocol

The experiments are conducted on the XM2VTS database [8], which contains video clips recorded on 295 subjects during four sessions taken at one month intervals. On each session, two frontal view and head rotation sequences were shot. The image data, which is included in the frontal view sequences, is used in the experiments.

The database obeys the Lausanne protocol [8]. The protocol specifies a partitioning of the database into three sets: a training set, an evaluation set, and a test set. The training set was used to build client models; the evaluation set was used to compute the decision (by estimating thresholds for instance, or parameters of a fusion algorithm); the test set was used only to estimate the performance.

The 295 subjects were divided into a set of 200 clients, 25 evaluation impostors, and 70 test impostors. Two different configurations were defined according to the protocol. They differ in the distribution of client training and client evaluation data.

All images are cropped and normalized to a size of $57 \times 61$, and are aligned based on the two eyes. In our framework, the position of the two eyes can be located either manually or automatically. In the experiments, the eyes are located manually. If the eyes are detected automatically, the verification accuracy of the respective methods considered will degrade due to the error in detecting the eyes.

### 4.2 Comparative results

In this section, we evaluate the performances of three face verification methods: (a) client specific fishface (CSF, also known as CS-LDA [2]); (b) $(2D)^2$ feature extraction using grey information ($(2D)^2$G); (c) $(2D)^2$ feature extraction using grey and skin color information ($(2D)^2$GC).

The comparison of the former two methods is to prove the performance of $(2D)^2$ feature extraction mainly; analogically, the compare between the last two methods is to verify that the skin color information is helpful for face verification.

For method $(2D)^2$GC, due to dealing with two features, a fusion decision --- Generalized Gradient Direction Metric (GGDM) [9, 10] is applied.

The thresholds in the decision making system have been determined based on the Equal Error Rate criterion, i.e. where the false rejection rate (FRR) is equal to the false acceptance rate (FAR).

In the experiments, to easy to compare, we unified the decision rules of the three methods. The rules are showed as follow:

$$\begin{cases} D \le T, \text{ reject claim} \\ D > T, \text{ accept claim} \end{cases} \quad (12)$$

where

$$D = |YM_c - YM_i| \quad (13)$$

and T is the equal error rate (ERR), where FAR and FRR are the same, was obtained on the evaluation set; Y is the projected feature matrix (Note: In CS-LDA, Y is a vector.); $M_c$ is the client mean matrix; $M_i$ is the imposter mean matrix.

The values in Table 1 on the test set provide the FAR, FRR and Total Error Rates (TER), i.e. the sum of false rejection and false acceptance rates.

The results are obvious in Table 1. Comparing CSF with $(2D)^2$G, the performance of the latter one is better than the former one. Because more information, which is useful for face verification, is extracted from the face image lengthways and breadthwise by $(2D)^2$ feature extraction. However, due to the own limit of FLD, the simple one dimension information can be obtained, which is deficient for face verification.

**Table 1** Comparison results of three Face verification methods on XM2VTS

| Method | Configuration I | | | Configuration II | | |
| --- | --- | --- | --- | --- | --- | --- |
| | FAR | FRR | TER | FAR | FRR | TER |
| CSF | 2.91 | 3.17 | 6.08 | 2.71 | 2.12 | 4.83 |
| $(2D)^2$G | 2.34 | 2.20 | 4.54 | 1.97 | 2.01 | 3.98 |
| $(2D)^2$GC | 1.77 | 1.64 | 3.41 | 1.31 | 1.47 | 2.78 |

In addition, the compare results of the last two methods have shown that the skin color distribution of the face can increase the performance of face verification.

4.3 Computation complexity

Because $(2D)^2$PCA and $(2D)^2$FLD are adopted in the method $(2D)^2$G to extract feature vector lengthways and breadthwise. So the computation complexity of $(2D)^2$G is higher than CSF whilst the PCA and FLD are taken as feature extraction method.

For the method $(2D)^2$GC, two kinds of feature information --- face grey and skin color information, are utilized. As a result, the computation complexity of $(2D)^2$GC almost increases twice compared with $(2D)^2$G approach.

As a result, the computation complexity of $(2D)^2$GC is highest among the three methods above. However, the computation complexities of the three methods are the same order. In addition, they all belong to the linear algorithms. So their computation complexities, which have little difference, are all low.

Contrast to the three methods above, ANN is a nonlinear algorithm; the parameters selection of SVMs is generally adjusted by genetic algorithm. So the computation complexities of ANN and SVMs are much higher than the former three methods.

# 5 Conclusions

In this paper, a new face verification algorithm $(2D)^2$GC is proposed. For the approach, on one hand, under the premise of guarantee of low computation complexity, high dimension feature extraction is adopted to improve the face verification accuracy; on the other hand, a new designed face skin color information, which is taken as an additive feature other than face grey feature, is added to the algorithm $(2D)^2$GC. The improvement is proved by the experiments in this paper. In addition, this paper also indicate that the method $(2D)^2$GC can be applied in some projects directly. Due to the features of $(2D)^2$GC summarized above, the algorithm can be implemented in the mobile telephone, PDA, smart card or stand-alone control access systems.

## References

1. Jacek Czyz, Josef Kittler, Luc Vandendorpe, Multiple classifier combination for face-based identity verification. Pattern Recognition, vol. 37, 2004, pp. 1459–1469.
2. Josef Kittler, Yongping Li, Jiri Matas, "Face verification using client specific Fisherfaces," in Proc. Int. Conf. Statistics of Directions, Shapes & Images, pp. 63--66, Sept. 2000.
3. S. Marcel and S. Bengio, Improving face verification using skin color information. In Proceedings of the 16th ICPR. IEEE Computer Society Press, 2002, pp. 378-381.
4. Daoqiang Zhang, Zhihua Zhou, $(2D)^2$PCA:2-Directional 2-Dimensional PCA for Efficient Face Representation and Recognition. Neurocomputing, vol. 69, 2005, pp. 224-231.
5. P. Nagabhushan, D.S. Guru, B.H. Shekar, $(2D)^2$FLD: An efficient approach for appearance based object recognition. Neurocomputing, vol. 69, 2006, pp. 934-940.
6. Chao Wang, Yongping Li and Xinyu Ao. Quality Fusion Rule for Face Recognition in Video. International Conference on Advanced Concepts for Intelligent Vision Systems, pp. 333-342, 2009.
7. S. Palanivel, B. Yegnanarayana, Multimodal person authentication using speech, face and visual speech. Computer Vision and Image Understanding, 2008, vol. 109, pp. 44-55.
8. J. Matas, M. Hamouz, K. Jonsson, J. Kittler, Y. Li, C. Kotropoulos, A. Tefas, I. Pitas, T. Tan, H. Yan, F. Smeraldi, J. Bigun, N. Capdevielle, W. Gerstner, S. Ben-Yacoub, Y. Abdeljaoued, and E. Mayoraz. Comparison of face verification results on the XM2VTS database. In A. Sanfeliu, J. J. Villanueva, M. Vanrell, R. Alqueraz, J. Crowley, and Y. Shirai, editors, Proceedings of the 15th ICPR, volume 4, pp. 858–863. IEEE Computer Society Press, 2000.
9. M. Sadeghi and J. Kittler, Decision making in the lda space: Generalised gradient direction metric. In The 6th Int. Conf. on Automatic Face and Gesture Recognition, Seoul, Korea, May, 2004.
10. J. Kittler,Yongping Li and J Matas, On matching scores for LDA-based face verification. The eleventh British Machine Vision Conference, 2000.